\pdfoutput=1

\documentclass[11pt]{article}

\usepackage{acl}

\usepackage{times}
\usepackage{latexsym}

\usepackage[T1]{fontenc}

\usepackage[utf8]{inputenc}

\usepackage{microtype}

\usepackage{inconsolata}

\usepackage{url}
\usepackage{subfigure}
\usepackage{multirow}
\usepackage{enumitem}
\usepackage{stmaryrd}
\usepackage{array}
\usepackage{threeparttable}
\usepackage{blindtext}
\usepackage{amssymb}

\usepackage[ruled,vlined,linesnumbered,longend]{algorithm2e}
\usepackage{svg}
\usepackage{placeins}

\usepackage{soul}
\usepackage[utf8]{inputenc}
\usepackage{graphicx}
\usepackage{amsmath}
\usepackage{booktabs}
\usepackage{lipsum}
\usepackage{listings}
\usepackage{xcolor, soul}
\sethlcolor{blue!10}

\usepackage{amsmath,amsfonts,bm}

\def\eqref#1{equation~\ref{#1}}

\def\1{\bm{1}}

\DeclareMathAlphabet{\mathsfit}{\encodingdefault}{\sfdefault}{m}{sl}
\SetMathAlphabet{\mathsfit}{bold}{\encodingdefault}{\sfdefault}{bx}{n}

\title{
An In-Context Schema Understanding Method \\for Knowledge Base Question Answering
}

\author{Yantao Liu,  Zixuan Li$^{*}$, Xiaolong Jin$^{*}$,Yucao Guo, Long Bai, Saiping Guan, \\
{\bf Jiafeng Guo} \and {\bf Xueqi Cheng} \\
        School of Computer Science and Technology, University of Chinese Academy of Sciences; \\ 
        CAS Key Laboratory of Network Data Science and Technology, \\
        Institute of Computing Technology, Chinese Academy of Sciences. \\
        \texttt{\{liuyantao22s, lizixuan, guoyuchan23b, jinxiaolong, bailong18b, guansaiping\}@ict.ac.cn} \\
        \texttt{\{guojiafeng, cxq\}@ict.ac.cn}
        }

\begin{document}
\maketitle
\begin{abstract}
    The Knowledge Base Question Answering (KBQA) task aims to answer natural language questions based on a given knowledge base. 
    Recently, Large Language Models (LLMs) have shown strong capabilities in language understanding and can be used to solve this task. 
    In doing so, a major challenge for LLMs is to overcome the immensity and heterogeneity of knowledge base schemas.
    Existing methods bypass this challenge by initially employing LLMs to generate drafts of logic forms without schema-specific details.
    Then, an extra module is used to inject schema information to these drafts.
    In contrast, in this paper, we propose a simple In-Context Schema Understanding (ICSU) method that enables LLMs to directly understand schemas by leveraging in-context learning. 
    Specifically, ICSU provides schema information to LLMs using schema-related annotated examples. 
    We investigate three example retrieval strategies based on raw questions, anonymized questions, and generated SPARQL queries. 
    Experimental results show that ICSU demonstrates competitive performance compared to baseline methods on both the KQA Pro and WebQSP datasets.
\end{abstract}

\section{Introduction}

The Knowledge Base Question Answering (KBQA) task, a challenging problem in the Natural Language Processing (NLP) field, focuses on understanding natural language questions and querying a given knowledge base (KB) to get answers. 
A kind of the common methods for this task is semantic parsing-based methods, where natural language questions are first converted into logical forms, such as SPARQL queries, and then executed on a given KB to retrieve answers.

\begin{figure}[tbp] 
    \centering
    \includegraphics[width=0.90\linewidth]{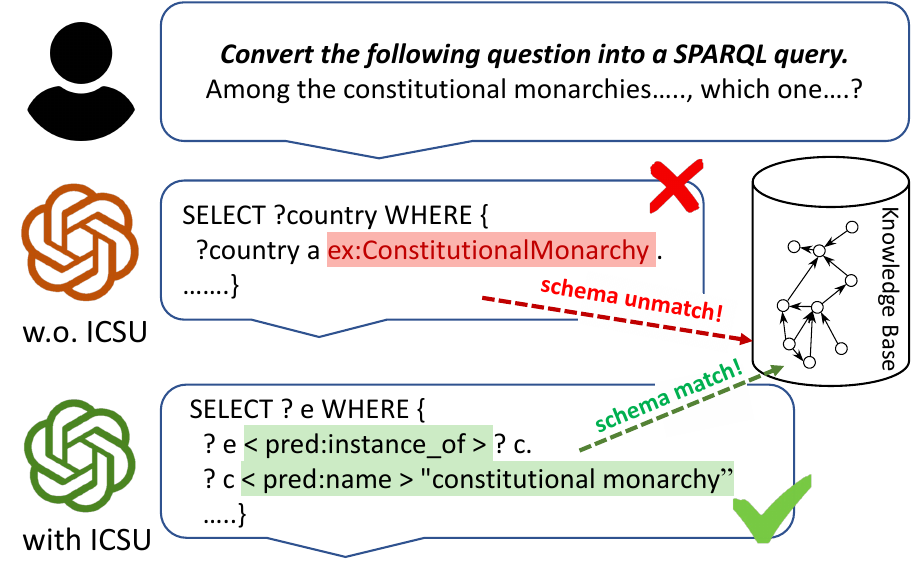}
    \caption{LLMs fail to generate a correct SPARQL query when lacking schema information.}
    \label{fig:intro}
    \vspace{-0.5cm}
\end{figure}

Recently, Large Language Models (LLMs) have shown impressive capabilities for generating formal languages~\cite{chen2021codex,nijkamp2022codegen,nijkamp2023codegen2}, suggesting that they can be adopted as semantic parsers in KBQA tasks.  
The key point for for LLMs in performing this task is to understanding the schema of the given KB.
For instance, LLMs need to understand the schema element ``<pred:instance\_of>'' and match it with the word ``Among'' in the question shown in Figure~\ref{fig:intro}.

However, understanding KB schemas presents challenges for LLMs in two aspects: 
\textbf{1) Heterogeneity}: The schema of a KB is user-defined and varies from one KB to another, making it difficult for LLMs to learn the schema information during pretraining.
\textbf{2) Immensity}: The number of schema elements can be enormous. 
For example, there are 8000+ relations in Wikidata~\cite{vrandevcic2014wikidata}.
This makes it impractical for LLMs to learn all schema elements on-the-fly by simply attaching the entire schema to the prompt.

Previous LLM-based KBQA methods~\cite{li-etal-2023-shot,nie2023code} attempt to bypass these challenges by only adopting LLMs to generate schema-free draft logical forms. 
Then, the schema information, such as relation IDs, is bound to the draft logical forms based on their surface names mentioned in the input questions by some retrieval methods (e.g., BM25 or FACC).
Finally, these logical forms are converted to SPARQL queries by external tools.
These methods rely on a complex, multi-stage pipeline and fail to fully exploit the abilities of LLMs.

Despite the heterogeneity and immensity of KB schemas, each question only relates to a few schema elements.
Notice that annotated question-SPARQL pairs contain schema information about corresponding questions,
we propose an In-Context Schema Understanding (ICSU) method to facilitate LLMs to directly generate SPARQL queries that match the KB schema by using a few annotated question-SPARQL pairs as examples in prompts. 
The key challenge here is retrieving appropriate examples containing comprehensive schema information related to the input questions. 
ICSU includes three example retrieval strategies based on raw questions, entity-anonymized questions, and draft SPARQL queries, respectively.
ICSU then adopts In-context Learning to prompt LLMs to generate SPARQL queries using examples retrieved through these strategies.
Compared with LLM-based baselines, ICSU achieves competitive performance on both KQA Pro and WebQSP datasets.

\section{The Proposed ICSU method}
\label{sec:method}
\begin{figure}[tbp] 
    \centering
    \includegraphics[width=\linewidth]{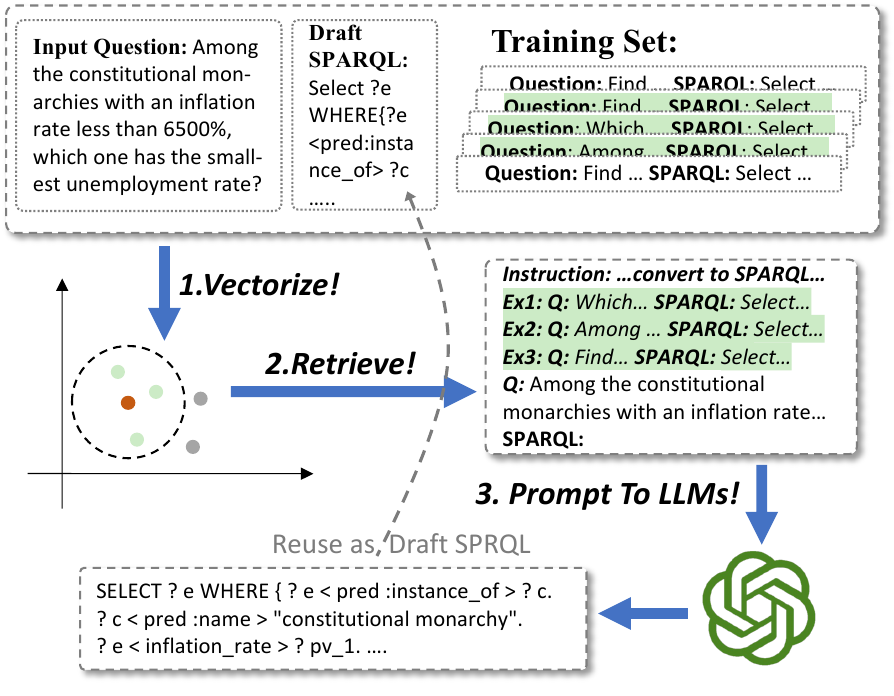}
    \caption{The pipeline of generating SPARQL queries with ICSU when example number $k$ = 3}
    \label{fig:pipeline}
    \vspace{-0.5cm}
\end{figure}

Figure~\ref{fig:pipeline} shows ICUS's pipeline.
1) ICSU vectorizes natural language questions or SPARQL queries for both the input and training set that has annotated question-SPARQL pairs.
2) ICSU retrieves examples from the training set based on the similarity between the vectors.
3) ICSU adopts In-context Learning to prompt LLMs to generate SPARQL queries based on the retrieved examples.
Note that the generated SPARQL query would be reused in the ICSU (SPARQL) as draft SPARQL query.

\subsection{ICSU Prompt for KBQA}
\label{sec:prompt}
To enable ICSU to generate accurate SPARQL queries, we design a prompt $x$ for each question. 
Specifically, $x=(i, \{e\}, q)$, contains three elements: 
1) A task instruction $i$ that provides a brief overview of the semantic parsing task in KBQA; 
2) An example set $\{e\}$ containing several examples that provide schema information for the given question; 
3) An input question $q$ that requires LLMs to provide the corresponding SPARQL query.

\subsection{Example Retrieval Strategies}
\label{all_strategies}
Here we introduce how to retrieve examples to construct the example set $\{e\}$ for the prompt $x$.

\paragraph{ICSU (Raw)}
\label{raw strategy}
is a raw-question based strategy.
Driven by the intuition that more similar examples can provide more related schema elements, 
ICSU computes similarities between the input question $q$ and questions in the training set that have annotated SPARQL queries.
It retrieves the top-k annotated question-SPARQL pairs with the highest similarity scores. 
Specifically, ICSU adopts MPNet~\cite{song2020mpnet}, a sentence embedding method, to get the embeddings of the questions and calculates the similarities based on negative Euclidean score. 

\paragraph{ICSU (Anonymized)}
\label{anonymize strategy}
is an anonymized-question based strategy.
Schema information is more about the relations between entities rather than specific entities. 
Besides, specific entities usually cannot be shared across different questions. 
Therefore, we propose to anonymize questions by replacing the entities with their corresponding entity types. 
Specifically, we use FLERT~\cite{schweter2020flert}, a NER model, to recognize types of entities. 
For different entities of the same type in the question, a number suffix is added for distinction.
For example, the question ``Which movie is shorter, The Greatest Story Ever Told or Rhinestone?'' 
will be transformed into "Which movie is shorter, [WORK\_OF\_ART\_0] or [WORK\_OF\_ART\_1]?”. 
The similarity between the anonymized questions is calculated in the same manner as ICSU (Raw).

\paragraph{ICSU (SPARQL)}
is a draft SPARQL based strategy.
The above two strategies are based on natural language questions but not
on SPARQL, which presents schema information more directly compared to natural language counterparts. 
Since the corresponding SPARQL of the input question is not available, we propose to generate a draft SPARQL query for the input question.
Specifically, we reuse the SPARQL queries generated by ICSU (Anonymized) as draft SPARQL queries.
Details about draft SPARQL generation can be found in Appendix~\ref{sec:appendix_draft_sparql}.
Then, we retrieve examples according to the similarity between the draft SPARQL queries and the annotated SPARQL queries from the training set. 
The similarity is calculated in the same manner as ICSU (Raw).

\paragraph{ICSU (Hybrid)}
Each of these three strategies has different retrieval preferences. 
To increase the diversity of the retrieved examples, we combine these three strategies to form a hybrid strategy. 
In particular, the order in which we combine the three strategies is as follows: the anonymized-question based strategy, the SPARQL based strategy, and the raw-question based strategy.

\paragraph{}
After acquiring the example set $\{e\}$ through retrieval strategies above,
we construct the prompt $x$ followed by the format in Section~\ref{sec:prompt} for input question $q$.
Finally, we  prompt LLMs to generate a SPARQL query based on the prompt $x$.

\section{Experiment}

\subsection{Datasets}

\textbf{KQA Pro}~\cite{cao2020kqa} is a large-scale dataset with MIT License for question answering over a subset of Wikidata~\cite{vrandevcic2014wikidata}, 
containing 94,376 and 11,797 question-SPARQL pairs in the training and test sets, respectively.
Although KQA Pro is built on a subset of Wikidata, its schema element is different.
For example, the relation “member of” is represented as \texttt{P102} in Wikidata, while it is presented as \texttt{<member\_of>} in KQA Pro’s KB.
This heterogeneity in schema prevents the information leakage that LLMs have been aware of KB schema at the pretraining stage.
Follow~\citet{cao2020kqa}, we report accuracy as each question in KQA Pro has only one answer.

\textbf{WebQSP}~\cite{yih-etal-2016-value} is a relatively small-scale dataset with CC-BY 4.0 License for question answering over Freebase~\cite{Bollacker2008FreebaseAC}. 
It contains 3,098 question-SPARQL pairs in the training set and 1,639 in the test set.
We use this dataset to investigate the effectivity of ICSU when annotated data is limited.
Follow~\citet{yih-etal-2015-stagg}, we report F1 score as one question in WebQSP have multiple answers.

\subsection{Experimental Setup}
Our main experiments are conducted with ChatGPT~\cite{ouyang2022training}, namely the \texttt{gpt-3.5-turbo} model through the OpenAI API.
The number of In-Context examples for ICSU is experimentally set to $6$. 
The decoding strategies is greedy search with temperature $t=0.7$.
We adopt KB-Binder~\cite{li-etal-2023-shot} and KB-Coder~\cite{nie2023code} as baselines.
Both baselines are based on ChatGPT and use BM25 as the entity/relation binder.
Additionally, we include ICSU with random examples to represent the performance of the normal In-Context Learning method.
In WebQSP, we report the results of KB-Binder and KB-Coder with 100-shot examples from the KB-Coder paper.
In KQA Pro, we reimplement KB-Binder with 6-shot examples and report the results.
Since KB-Coder is not open-source, its result is not available.

\subsection{Experimental Results}

\begin{table}
    \centering
    \scalebox{0.8}{
    \begin{tabular}{lcc} 
        \toprule
        \textbf{Methods} & \textbf{KQA Pro} & \textbf{WebQSP} \\ 
        \midrule
        ICSU \small(Random) & 5.01 & 17.00 \\
        KB-BINDER (1)-R & 69.81 & 68.90 \\
        KB-Coder (1)-R & - & 72.20 \\
        \midrule
        ICSU \small(Raw) & 68.97 & 59.03 \\
        ICSU \small(Anonymized) & 73.32 & 61.73 \\
        ICSU \small(SPARQL) & 73.66 & \textbf{72.36} \\
        ICSU \small(Hybrid) & \textbf{76.16} & 69.23 \\
        \bottomrule        
    \end{tabular}
    }
    \caption{
        Comparison with KB-BINDER and KB-Coder on KQA Pro in Accuracy (\%) and WebQSP in F1 (\%).
    }
    \label{tab:baseline}
    \vspace{-0.5cm}
\end{table}
All four strategies significantly outperform ICSU (Random) in both datasets, indicating our method works whether annotated data is sufficient or limited.
ICSU (Anonymized) provides satisfactory improvement compared to ICSU (Raw).
This result testifies that the schema information is more about the relations rather than the entities.
Furthermore, ICSU (SPARQL) achieves better performance compared to ICSU (Anonymized) in both datasets.
This shows that formal language queries can present more precise schema information compared to their natural language counterparts when used to retrieve examples.
In KQA Pro, with the same number of examples, all four strategies present competitive performance compared to the KB-Binder baseline.
In WebQSP, 
although the performance of KB-Binder and KB-Coder is achieved with 100-shot examples, 
the performance of ICSU (SPARQL) with 6-shot examples is comparable.

\subsection{ICSU with Different LLMs}
\begin{table}[t]
    \scalebox{0.75}{
    \setlength{\tabcolsep}{3pt}
    \renewcommand{\arraystretch}{1.2}
    \centering

        \begin{tabular}{llcccc}
        \toprule
        & \textbf{Methods} & \textbf{LLaMA} & \textbf{Alpaca} & \textbf{ChatGPT} & \textbf{InstructGPT} \\ 
        \midrule
        \multirow{4}{*}{\rotatebox{90}{\textbf{KQA Pro}}}
        & ICSU \small{(Raw)}        & 14.90             &       40.80        & 68.97            & 71.58                     \\
        & ICSU \small{(Anonymized)}  & 20.50             &       52.27        & 73.32            & 75.32                     \\
        & ICSU \small{(SPARQL)}     & \textbf{23.50}    &\textbf{54.82}      & 73.66            & 76.37                     \\
        & ICSU \small{(Hybrid)}      & 21.51             &       52.88        & \textbf{76.16}   & \textbf{78.76}            \\ 
        \midrule
        \multirow{4}{*}{\rotatebox{90}{\textbf{WebQSP}}}
        & ICSU \small{(Raw)}               & 4.75             & 35.66             & 59.03           & 62.54                   \\
        & ICSU \small{(Anonymized)}        & 5.16             & 42.61             & 61.73           & 63.24                   \\
        & ICSU \small{(SPARQL)}            & \textbf{6.23}    & \textbf{47.78}    & \textbf{72.36}  & \textbf{74.99}          \\
        & ICSU \small{(Hybrid)}            & 6.19             & 42.92             & 69.23           & 74.12                   \\ \bottomrule
        \end{tabular}
    }
    \caption{
        ICSU Performance with Different LLMs: Accuracy (\%) on KQA Pro and F1 Score (\%) on WebQSP.
    }
    \label{tab:icsu_llm}
    \vspace{-0.5cm}
\end{table}

One interesting observation is that the performance of ICSU (Hybrid) is not always better than ICSU (SPARQL).
To investigate this thoroughly, we test the performance of ICSU with different LLMs in both datasets.
Specifically, we adopt LLaMA-7B~\cite{touvron2023llama} and Alpaca-7B~\cite{taori2023alpaca} ChatGPT and InstructGPT~\cite{ouyang2022training}.
Note InstructGPT is acquired from \texttt{text-davinci-003} API from OpenAI.

The results are shown in Table~\ref{tab:icsu_llm}.
In KQA Pro, ICSU (Hybrid) gets the best performance with ChatGPT and InstructGPT, while it underperforms ICSU (SPARQL) with LLaMA-7B and Alpaca-7B.
In WebQSP, ICSU (Hybrid) underperforms ICSU (SPARQL) with all tested LLMs.

The key of ICSU is to retrieve input-question-related schema elements from annotated question-SPARQL pairs.
ICSU (Hybrid) increases the recall rate of schema elements by increasing the diversity of examples, but at the same time, the precision of schema elements is decreased.
Powerful LLMs, like ChatGPT or InstructGPT, achieve the best results due to their abilities to benefit from the increased recall rate and not be affected by the decreased precision.
However, LLaMA-7B and Alpaca-7B are not as capable and suffer from the decreased precision.
Thus, in KQA Pro, ICSU (Hybrid) surpasses ICSU (SPARQL) with ChatGPT and InstructGPT.
In WebQSP, 98\% questions are one-hop, limiting its related schema elements (relation).
Therefore, increasing the diversity of examples does not benefit the recall rate but harms the precision rate of retrieved schema elements, making ICSU (Hybrid) worse than ICSU (SPARQL).

\subsection{Effect of Schema Recall Rate}

\begin{figure}[t]
    \centering
    \includegraphics[width=0.8\linewidth]{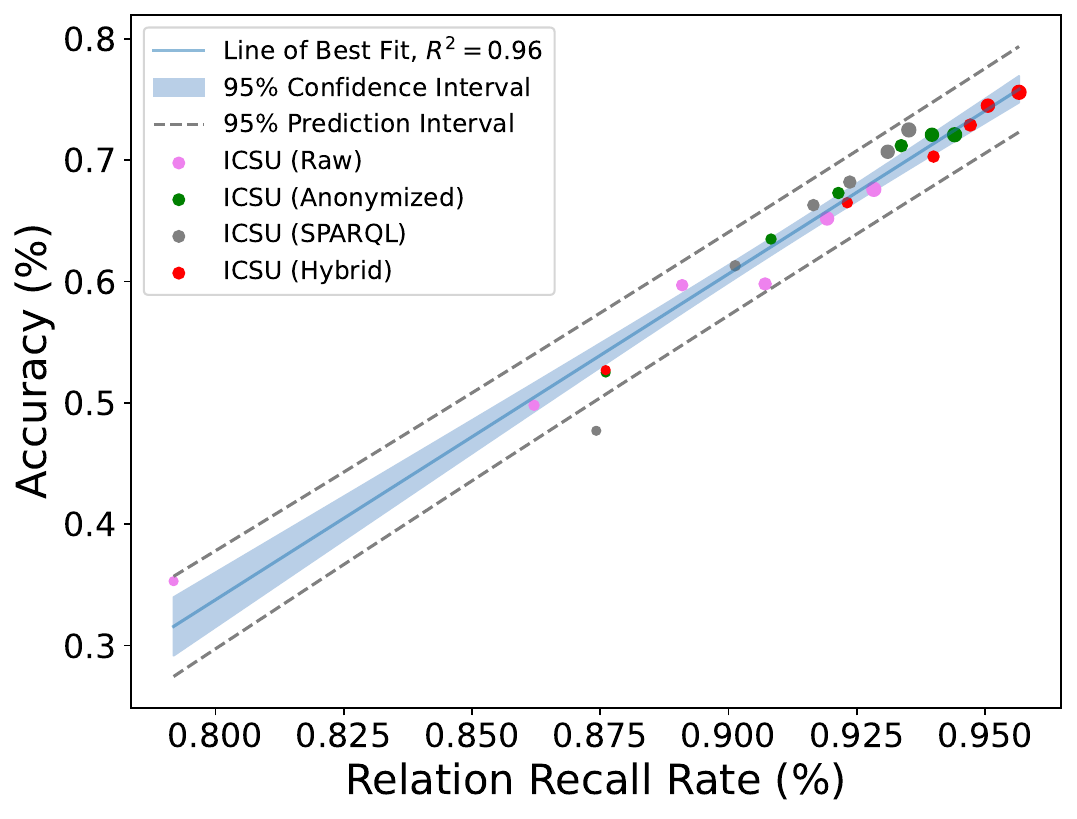}
\caption{The correlation between relation recall rate and accuracy on KQA Pro.}
\label{fig:cover2acc}
\vspace{-0.5cm}
\end{figure}

To study how schema information contributes to the final results, we conduct a detailed analysis of ICSU with ChatGPT on the validation set of KQA Pro. 
We use the relation recall rate to reflect the schema elements' recall rate in the retrieved examples.
Specifically, we conduct statistics on (relation recall, accuracy) pairs of the ICSU with four proposed retrieval strategies under $k (k=1,...,6)$ examples and plot them on Figure \ref{fig:cover2acc}. 
It can be observed that the accuracy increases when the relation recall rate gets larger. 
This result verifies that the key to the success of ICSU is to retrieve examples that contain comprehensive schema information related to the input questions.

\section{Related Work}

Semantic parsing methods~\cite{yih-etal-2015-stagg, Wu2021FromPT, shin2021constrained,xu2020autoqa}, as common approaches in KBQA, convert natural language questions into formal language queries for execution on knowledge bases, offering precise answers and interpretability.

Recently, LLMs such as GPT-$3$ \cite{brown2020language} and LLaMA~\cite{touvron2023llama} have become noteworthy in KBQA due to their impressive performance on formal language generation tasks.
Particularly, in KBQA, 
\citet{tan2023evaluation} evaluates the performance of LLMs in KBQA tasks via directly using LLMs as knowledge bases. 
\citet{baek2023knowledge} searches the relevant facts and attaches them to the input question from the knowledge base to enhance the ability of LLMs on KBQA.
\citet{li-etal-2023-shot} and \citet{nie2023code} leverage LLMs to generate a draft logical form and then inject schema information into it by some external tools.

\section{Conclusion}
In this paper, we propose the In-Context Schema Understanding (ICSU) method for enabling LLMs to directly generate SPARQL queries for KBQA.
ICSU adopts the In-context Learning to instruct LLMs to generate SPARQL queries with examples retrieved via four different strategies.
Experimental results show that ICSU achieves competitive performance compared to state-of-the-art LLM-based methods on both KQA Pro and WebQSP datasets.

\section * {Limitations}
In order to answer a question with a knowledge base, an entity linking process is required to map the entities in the question to the entities in the knowledge base.
However, our study focus on the semantic parsing of KBQA, and our method assumes that we have the linked entities.
Our method will be undoubtedly affected by the error from entity process, and are not able to recover the error in the entity linking process.
More details about entity linking can be viewed in appendix~\ref{sec:entitylinking}.

\section * {Ethical Considerations}
Our work focus on generating SPARQL queries with LLMs for KBQA.
ICSU relies on LLMs from open-source community and OpenAI API.
We strictly follows the terms of service of OpenAI API and open-source licenses.
Our work is based on the public datasets, and we do not collect any data from human subjects.

\bibliography{1-ref}
\bibliographystyle{acl_natbib}

\clearpage

\appendix

\label{sec:appendix}
\section{Appendix}

\subsection{Entity Linking Details}
\label{sec:entitylinking}
In order to answer a question with a knowledge base, an entity linking process is required to map the entities in the question to the entities in the knowledge base.

Particularly, In KQA Pro dataset, entity linking is achieved by applying a relation constraint in the SPARQL query. 
For example, the question "Where was Sabrina (which has Angie Dickinson as a cast member) published on 1996-01-11?", the entities are appropriately linked using the SPARQL relations ?e <pred:name> "Sabrina" and ?e\_1 <pred:name> "Angie Dickinson". 
Namely, in KQA Pro, the entity linking is achieved by using the pred:name relation in the SPARQL query to find one entity that matches the entity name in the question.

As for WebQSP Dataset, we followed the previous work ReaRev~\cite{mavromatis-karypis-2022-rearev} using the seed entities provided by \citet{yih-etal-2016-value} for WebQSP. Then We add "Topic Entity Id: xxxx" in the prompt after the natural question input, which effectively signals the LLMs about the topic entity id information. 

It is crucial to note that our contributions are on the semantic parsing part, and our method assumes that we have the linked entities. Our approach does not exclusively hinge on a specific entity linking method. Alternative methods like BLINK~\cite{wu-etal-2020-scalable} or ELQ~\cite{li-etal-2020-efficient} can be easily integrated and employed if deemed appropriate.

\subsection{The Effect of the Number of Examples $k$}

We sampled 1,000 samples from the development dataset of KQA Pro to analyze performance when varying the number of in-context learning examples from 0 to 8. 
Table~\ref{tab:effect_of_k} shows the results.
\begin{table}[h]
    \centering
    \scalebox{0.8}{
        \begin{tabular}{lccccc}
        \toprule
        Acc over $k$ & 0 & 2 & 4 & 6 & 8 \\
        \midrule
        ICSU \small(Raw) & 0.0 & 49.8 & 59.8 & 67.6 & 70.1 \\
        ICSU \small(Anonymized) & 0.0 & 63.5 & 71.2 & 72.1 & 73.3 \\
        ICSU \small(SPARQL) & 0.0 & 61.3 & 68.2 & 72.5 & 73.5 \\
        ICSU \small(Hybrid) & 0.0 & 66.5 & 72.9 & 75.6 & 76.7 \\
        \bottomrule
        \end{tabular}
    }
    \caption{The effect of the number of examples $k$ to ICSU on accuracy in KQA Pro.}
    \label{tab:effect_of_k}
\end{table}
It can be observed that the performance of ICSU increases along with the shot number increases. And it tends to converge after a shot number of $k=6$. 
Considering the balance between performance and computational cost, we finally chose $6$ as the number of examples for in-context learning in our main experiment on KQA Pro.

Another interesting observation is that the performance of ICSU (SPARQL) is worse than ICSU (Anonymized) when $k=2$ or $k=4$.
The motivation of ICSU (SPARQL) is polishing, namely, the draft SPARQL queries in ICSU (SPARQL) are based on the generated results of ICSU (Anonymized).
Therefore, errors from ICSU (Anonymized) will undoubtedly affect the accuracy of SPARQL retrieval.
Particularly, the number of examples provided to the ICSU (Anonymized) influences the quality of the draft SPARQL queries it generates, which further impacts the retrieval of ICSU (SPARQL). 
When the sample size of the ICSU (Anonymized) is limited (only 2 or 4), the quality of the generated draft SPARQL queries is relatively poor, introducing noise that could adversely affect the performance of ICSU (SPARQL).

\subsection{More Experiment Results}

Table~\ref{tab:webqsp_em} presents the exactly match score of ICSU on WebQSP with four different LLMs.
It can be observed the result in Exactly Match (EM) is consistent to the F1 score in Table~\ref{tab:icsu_llm}.
\begin{table}[!h]
    \setlength{\tabcolsep}{3pt}
    \renewcommand{\arraystretch}{1.2}
    \centering

    \scalebox{0.8}{
        \begin{tabular}{@{}lccccc@{}}
        \toprule
        \textbf{Methods} & \textbf{LLaMA} & \textbf{Alpaca} & \textbf{ChatGPT} & \textbf{InstructGPT} \\ \midrule
        \textbf{ICSU \small{(w.o Ex)}}     & 0.00             & 0.00              & 0.00            & 0.00                    \\
        \textbf{ICSU \small{(Random)}}      & 0.79             & 5.49              & 15.44           & 13.91                   \\
        \midrule
        \textbf{ICSU \small{(Raw)}}         & 4.33             & 33.19             & 55.52           & 59.73                   \\
        \textbf{ICSU \small{(Anonymized)}}  & 4.58             & 40.02             & 58.76           & 60.71                   \\
        \textbf{ICSU \small{(SPARQL)}}      & \textbf{5.80}    & \textbf{46.92}    & \textbf{70.35}  & \textbf{73.22}                   \\
        \textbf{ICSU \small{(Hybrid)}}      & 5.49             & 40.51             & 65.89           & 71.32                   \\ \bottomrule
        \end{tabular}
    }
    \caption{
        ICSU results on WebQSP on EM (\%).
    }
    \label{tab:webqsp_em}
\end{table}

Table~\ref{tab:kqapro-detailed-result} presents the detailed results of ICSU and baselines on KQA Pro.
It can be observed that in all sorts of questions, ICSU demonstrate a competitive performance compared to baseline.


\begin{table*}[htbp]
    \centering
\begin{tabular}{@{}lcccccccc@{}}
\toprule
\textbf{Methods} & \textbf{Overall} & \textbf{Multihop} & \textbf{Qualifier} & \textbf{Comparison} & \textbf{Logical} & \textbf{Count} & \textbf{Verify} \\ 
\midrule
KB-BINDER (1)-R                   & 69.81            & 64.54             & 58.38              & 86.39               & 60.70            & 66.67           & 77.76   \\
\midrule
ICSU \small{(Raw)}                 & 68.97            & 63.49             & 57.32              & 85.55               & 58.93            & 66.06          & 77.83   \\
ICSU \small{(Anonymized)}           & 73.32            & 68.28             & 62.92              & 89.10                & 65.08            & 66.97          & 78.66  \\
ICSU \small{(SPARQL)}              & 73.66            & 68.97             & 62.71              & 88.91               & 65.08            & 65.84          & 80.39   \\
ICSU \small{(Hybrid)}               & \textbf{76.16}            & \textbf{72.11}   & \textbf{65.72}              & \textbf{90.46}               & \textbf{69.30}            & \textbf{68.25}          & \textbf{83.22}  \\ 
\bottomrule
\end{tabular}
\caption{Detailed accuracy (\%) results of ICSU and KB-Binder on KQA Pro with ChatGPT}
\label{tab:kqapro-detailed-result}
\end{table*}

\subsection{Details for Generating a Draft SPARQL Query}
\label{sec:appendix_draft_sparql}
The intuition of ICSU (SPARQL) is to polish the SPARQL generated by other strategies, which is called "draft SPARQL query” in ICSU (SPARQL). The whole process of ICSU (SPARQL) is as follows. First, we use other strategies like ICSU (Anonymized) to generate a SPARQL query as a draft SPARQL query. Second, we retrieve examples again by comparing their grounded SPARQL queries with this draft SPARQL query according to the text similarity. Finally, these following retrieved examples are utilized in ICSU (SPARQL) to generate the final SPARQL query, which is then executed to obtain the final answer.

Thus, the detailed process of generating draft SPARQL query is the same as ICSU (Anonymized). Specifically, we list the process as follows:
\textbf{1)} We first retrieve six examples from the training dataset by comparing the text-similarity bewteen anonymized questions.
\textbf{2)} We then utilize these six examples to construct a prompt in the following format:
For instance, if the question is `Where was Sabrina (which has Angie Dickinson as a cast member) published on 1996-01-11?', the final prompt would be:

\begin{table}[h]
    \centering
    \scalebox{0.72}{
        \begin{tabular}{p{1.3\linewidth}}
            \toprule
                \vspace{-2mm}
                \textbf{Instruction}: You are given natural questions that could be answered over some complex reasoning steps on one knowledge base. Your task is to convert the given natural questions into SPARQL queries which can be excuted to find out the answer. \\
                Input 1: Retrieved example 1's question \\
                Output 1: Retrieved example 1's SPARQL query \\
                Input 2: Retrieved example 2's question \\
                Output 2: Retrieved example 2's SPARQL query \\
                Input 3: Retrieved example 3's question \\
                Output 3: Retrieved example 3's SPARQL query \\
                Input 4: Retrieved example 4's question \\
                Output 4: Retrieved example 4's SPARQL query \\
                Input 5: Retrieved example 5's question \\
                Output 5: Retrieved example 5's SPARQL query \\
                Input 6: Retrieved example 6's question \\
                Output 6: Retrieved example 6's SPARQL query \\
                Input 7: Where was Sabrina (which has Angie Dickinson as a cast member) published on 1996-01-11? \\
                Output 7: \\
            \bottomrule
        \end{tabular}
    }
    \caption{
        The prompt of for generating draft SPARQL query.
    }
    \label{tab:draft_sparql_prompt}
\end{table}
\textbf{3)} Finally, we feed this prompt to LLMs to generate the draft SPARQL query.

\subsection{Schema Information Leakage}
\label{sec:appendix_schema_leakage}.
Here we analyze for some common knowledge bases like Wikidata or Freebase, whether the schema information is leaked to LLMs during the pretraining stage.
We conduct the following experiments to verify this.
We provide one property (relation) and its corresponding description in KB to LLMs as the example, and then we ask LLMs to generate more property-description pairs.
As shown in Table~\ref{tab:freebase_leakage} and Table~\ref{tab:wikidata_leakage}, LLMs can generate more property-description pairs with high quality.
After carefully human check, all the property and their corresponding descriptions are correct and consistent with the knowledge base. 
The result indicates that LLMs have already been aware of the schema information of the common knowledge bases during the pretraining stage.
In this case, existing KBQA datasets whose schema is identical to the widely used knowledge bases (e.g., Wikidata, Freebase) are not suitable for evaluating our method,
since the schema information is already leaked to LLMs during the pretraining stage.

However, KQA Pro is built on a subset of Wikidata, and its schema element is different.
This heterogeneity in schema prevents the information leakage that LLMs have been aware of KB schema at the pretraining stage.
As shown in Table~\ref{tab:kqapro_leakage}, LLMs do not know KQA Pro and its schema information.
Therefore, KQA Pro is suitable for evaluating our method.

\begin{table}[h]
    \centering
    \scalebox{0.72}{
        \begin{tabular}{p{1.3\linewidth}}
            \toprule
                \vspace{-2mm}
            \textbf{\textsc{User:}} Please give out some property in KQA Pro and explain the meaning of them \\
            For example: \\
            Property: <pred:instance\_of> \\
            Description: that class of which this subject is a particular example and member \\
            \midrule
            \textbf{\textsc{ChatGPT:}} I'm sorry, but as of my last knowledge update in January 2022, I don't have specific information about a knowledge graph or database called "KQA Pro" or its properties. It's possible that "KQA Pro" is a proprietary or specialized knowledge base that was created or updated after my last update. \\
            \\
            If you have a specific question about a knowledge graph or property, please provide more context or details, and I'll do my best to provide a meaningful response based on the information I have up to my last update. \\
            \bottomrule
        \end{tabular}
    }
    \caption{
        ChatGPT does not know KQA Pro.
    }
    \label{tab:kqapro_leakage}
\end{table}

\begin{table}[h]
    \centering
    \scalebox{0.72}{
        \begin{tabular}{p{1.3\linewidth}}
            \toprule
                \vspace{-2mm}
            \textbf{\textsc{User:}} plz give out some property in freebase and explain the meaning of them \\
            for example: \\
            Property: ns:government.politician.government\_positions\_held \\
            Dataescription: Government positions held by this politician. \\
            \midrule
        
            \textbf{\textsc{ChatGPT:}} Certainly! Freebase was a collaborative knowledge base that was maintained by the community and later incorporated into Google's Knowledge Graph. It contained a vast amount of structured information about a wide range of topics. Here are a few more properties from Freebase along with their meanings: \\
            \\
            Property: ns:common.topic.notable\_types \\
            Description: This property was used to identify the notable types or categories that a given topic belonged to. For example, it could be used to specify that a person is notable as an actor, musician, scientist, or politician. \\
            \\
            Property: ns:people.person.place\_of\_birth \\
            Description: This property indicated the place where a person was born. It provided information about a notable person's place of birth, which could be a city, town, or country. \\
            \\
            Property: ns:location.location.geolocation \\
            Description: Geolocation properties were used to specify the latitude and longitude coordinates of a location. This allowed for the mapping and pinpointing of places on the Earth's surface. \\
            \\
            Property: ns:book.author \\
            Description: This property was used to link an author to a book or literary work. It indicated that a particular person was the author of a specific book, novel, or other written material. \\
            \\
            Property: ns:film.film.directed\_by \\
            Description: This property linked a film to its director. It indicated the person who directed a particular film, providing information about the film's directorial credits. \\
            \\
            Property: ns:organization.organization.founders \\
            Description: This property listed the founders or co-founders of an organization. It helped establish the individuals responsible for creating or establishing a particular company or institution. \\
            \\
            Property: ns:event.event.date \\
            Description: Event date properties were used to specify the date of occurrence of an event. This could be an important historical event, a festival, a concert, or any other significant happening. \\
            \\
            Property: ns:business.business\_operation.industry \\
            Description: This property indicated the industry or sector in which a business operated. It helped categorize businesses based on their primary field of activity, such as technology, healthcare, or finance. \\
            \\
            Property: ns:film.film.genre \\
            Description: Genre properties were used to categorize films based on their genre or style. Examples of film genres include action, comedy, drama, science fiction, and horror. \\
            \\
            Property: ns:book.book.subjects \\
            Description: This property listed the subjects or topics covered by a book. It provided information about the main themes or subject matter of a particular literary work. \\
            \bottomrule
        \end{tabular}
    }
    \caption{
        The example of schema information leakage in ChatGPT on Freebase.
    }
    \label{tab:freebase_leakage}
\end{table}

\begin{table}[h]
    \centering
    \scalebox{0.72}{
        \begin{tabular}{p{1.3\linewidth}}
            \toprule
                \vspace{-2mm}
            \textbf{\textsc{User:}} Please give out some property in Wikidata and explain the meaning of them \\
            For example: \\
            Property: P108 (employer) \\
            Description: Person or organization for which the subject works or worked \\
            \midrule
            \textbf{\textsc{ChatGPT:}} Certainly! Here are some properties from Wikidata along with their descriptions: \\
            \\
            Property: P31 (instance of) \\
            Description: This property indicates the type or class of an entity. It is used to classify items into broader categories. \\
            \\
            Property: P569 (date of birth) \\
            Description: This property represents the date on which a person was born. \\
            \\
            Property: P570 (date of death) \\
            Description: This property represents the date on which a person or entity died or was dissolved. \\
            \\
            Property: P21 (sex or gender) \\
            Description: This property specifies the gender or sex of a person. It can be used to indicate whether a person is male, female, or has another gender identity. \\
            \\
            Property: P625 (coordinate location) \\
            Description: This property provides the geographical coordinates (latitude and longitude) of a specific location on Earth. \\
            \\
            Property: P131 (located in the administrative territorial entity) \\
            Description: This property links a location to its administrative or political division, such as a city within a country or a district within a state. \\
            \\
            Property: P1082 (population) \\
            Description: This property represents the estimated or recorded population of a place or region at a specific point in time. \\
            \\
            Property: P17 (country) \\
            Description: This property links an entity to the country it is primarily associated with or located in. \\
            \\
            Property: P36 (capital) \\
            Description: This property specifies the capital city of a country or administrative division. \\
            \\
            Property: P47 (shares border with) \\
            Description: This property is used to link geographical entities that share a border or boundary with each other. \\
            \\
            Property: P106 (occupation) \\
            Description: This property specifies the occupation or profession of a person. \\
            \\
            Property: P127 (owned by) \\
            Description: This property links an item to the owner or entity that possesses it. \\
            \\
            Property: P279 (subclass of) \\
            Description: This property is used to indicate that one item is a subclass or more specific instance of another item. \\

            \bottomrule
        \end{tabular}
    }
    \caption{
        The example of schema information leakage in ChatGPT on Wikidata.
    }
    \label{tab:wikidata_leakage}
\end{table}

\end{document}